\documentclass[sigconf]{acmart}

\usepackage{booktabs} 

\usepackage{graphicx}
\usepackage{subcaption}
\usepackage[utf8]{inputenc}
\usepackage[export]{adjustbox}
\usepackage{wrapfig}

\usepackage{array} 
\usepackage{multirow} 
\usepackage{xspace} 
\usepackage{balance}

\setcopyright{rightsretained}

\newcommand{\myData}{StaQC\xspace}
\newcommand{\nop}[1]{}
\newcommand{\codeblock}{\textsc{CodeBlock}\xspace}

\newcommand{\from}[2]{\textcolor{blue}{{\bf [{\sc from #1:} #2]}}}

\begin{document}

\copyrightyear{2018}
\acmYear{2018} 
\setcopyright{iw3c2w3}
\acmConference[WWW 2018]{The 2018 Web Conference}{April 23--27, 2018}{Lyon, France}
\acmBooktitle{WWW 2018: The 2018 Web Conference, April 23--27, 2018, Lyon, France}
\acmPrice{}
\acmDOI{10.1145/3178876.3186081}
\acmISBN{978-1-4503-5639-8/18/04}

\fancyhead{}

\title{\myData: A Systematically Mined Question-Code Dataset \\from Stack Overflow}

\author{
    Ziyu Yao$^\dagger$, Daniel S. Weld$^\#$, Wei-Peng Chen$^{\dagger\dagger}$, Huan Sun$^\dagger$
} 
\affiliation{
  \institution{$^\dagger$The Ohio State University, $^\#$University of Washington, $^{\dagger\dagger}$Fujitsu Labs of America}
}
\email{{yao.470, sun.397}@osu.edu, weld@cs.washington.edu, wei-peng.chen@us.fujitsu.com}


\begin{abstract}
Stack Overflow (SO) has been a great source of natural language questions and their code solutions (i.e., question-code pairs), which are critical for many tasks including code retrieval and annotation. In most existing research, question-code pairs were collected \textit{heuristically} and tend to have low quality.
In this paper, we investigate a new problem of \textit{systematically mining} question-code pairs from Stack Overflow (in contrast to \textit{heuristically collecting} them).
It is formulated as predicting whether or not a code snippet is a standalone solution to a question.
We propose a novel Bi-View Hierarchical Neural Network which can capture both the programming content and the textual context of a code snippet (i.e., two views) to make a prediction. On two manually annotated datasets in Python and SQL domain, our framework substantially outperforms heuristic methods with at least \textbf{15\%} higher $F_1$ and accuracy.
Furthermore, we present \myData (\underline{Sta}ck Overflow \underline{Q}uestion-\underline{C}ode pairs), the largest dataset to date of \textbf{$\sim$148K} Python and \textbf{$\sim$120K} SQL question-code pairs, automatically mined from SO using our framework. Under various case studies, we demonstrate that \myData can greatly help develop data-hungry models for associating natural language with programming language\footnote{Available at \url{https://github.com/LittleYUYU/StackOverflow-Question-Code-Dataset}.}.
\end{abstract}

%
%



\maketitle

\section{Introduction} \label{introduction}
Online forums such as Stack Overflow (SO) \cite{stackoverflow} have contributed a huge number of code snippets, understanding and reuse of which can greatly speed up software development. Towards this goal, a lot of research work have been developed recently, such as retrieving or generating code snippets based on a natural language query, and annotating code snippets using natural language \cite{allamanis2015bimodal, iyer1summarizing, zilberstein2016leveraging, yin17acl, rabinovich2017abstract, loyola2017neural, building-natural-language-interfaces-web-apis}. At the core of these work are machine learning models that map between natural language and programming language, which are typically data-hungry \cite{NIPS2012_4824, ratner2016data, Goodfellow-et-al-2016} and require large-scale and high-quality $<$natural language question, code solution$>$ pairs (i.e., question-code or QC pairs).   

\begin{figure}[t!]
\begin{center}
\includegraphics[width=0.48\textwidth]{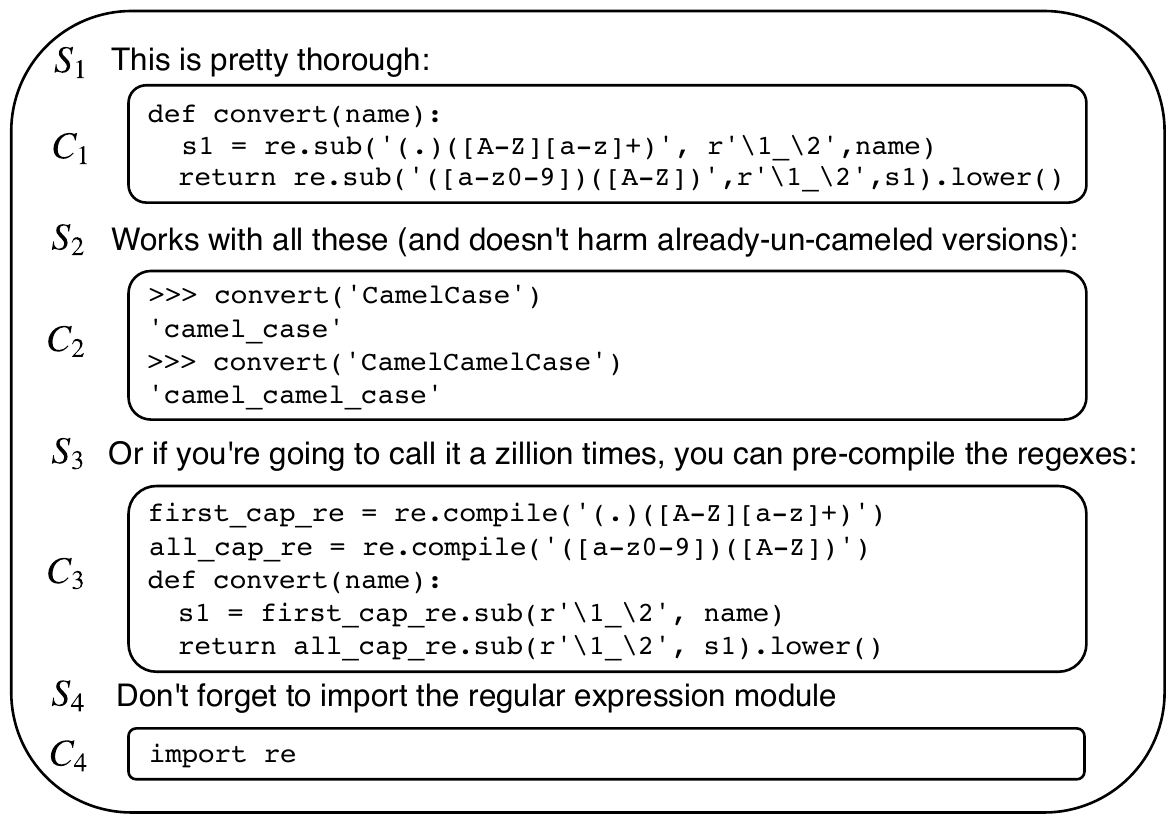}
\caption{The accepted answer post to question \textit{``Elegant Python function to convert CamelCase to snake\_case?''} in SO. $S_i$ ($i=1, 2, 3, 4$) and $C_j$ ($j=1, 2, 3, 4$) denote sentence blocks and code blocks respectively, which can be trivially separated based on the HTML format.}
\label{fig:runningexample}
\end{center}
\end{figure}

In our work, \textit{we define a code snippet as a code solution when the questioner can solve the problem solely based on it} (also named as ``\text{standalone}'' solution). Take Figure \ref{fig:runningexample} as an example, which shows the \text{accepted} answer post\footnote{In SO, an accepted answer post is marked with a green check by the questioner, if he/she thinks it solves the problem. {Following previous work \cite{iyer1summarizing, yang2016query},} although there can be multiple answer posts to a question, \textit{we only consider the accepted one because of its verified quality}, and use ``accepted answer post'' and ``answer post'' interchangeably.} to question ``\emph{Elegant Python function to convert CamelCase to snake\_case}''. Among the four code snippets \{$C_1$, $C_2$, $C_3$, $C_4$\}, only $C_1$ and $C_3$ are standalone code solutions to the question while the rest are not, because $C_2$ only gives an input-output demo of the ``convert'' function without its definition and $C_4$ is a reminder of an additional detail. Given an answer post with multiple code snippets (i.e., a multi-code answer post) like Figure \ref{fig:runningexample}, previous work usually collected question-code pairs in heuristic ways: Simply pair the question title with the first code snippet, or with each code snippet, or with the concatenation of all code snippets in the post \cite{allamanis2015bimodal, zilberstein2016leveraging}. Iyer et al. \cite{iyer1summarizing} merely employed accepted answer posts that contain exactly one code snippet, and discarded all others with multiple code snippets.
Such heuristic question-code collection methods suffer from at least one of the following weaknesses: (1) Low precision: Questions do not match with their paired code snippets, when the latter serve as background, explanation, or input-output demo rather than as a solution (e.g., $C_2$ in Figure \ref{fig:runningexample}); (2) Low recall: If one only selects the first code snippet to pair with a question, other code solutions in an answer post (e.g., $C_3$) will be unemployed. 

In fact, multi-code answer posts are very common in SO, which makes the low-precision and low-recall issues even more prominent. In the Stack Exchange Data dump\cite{StackExchangeDataDump}, among all accepted answer posts for Python and SQL ``how-to-do-it'' questions (to be introduced in Section \ref{preliminaries}), 44.66\% and 34.35\% contain more than one code snippets respectively.
Note that an accepted answer post was verified only \textit{as an entirety} by the questioner, and labels on whether each individual code snippet serves as a standalone solution or not are not readily available. Moreover, it is not feasible to obtain such labels by simply running each code snippet in a programming environment for two reasons: (1) A runnable code snippet is not necessarily a code solution (e.g., $C_4$ in Figure~\ref{fig:runningexample}); (2) It was reported that around 74\% of Python and 88\% of SQL code snippets in SO are not directly parsable or runnable \cite{iyer1summarizing, yang2016query}. {Nevertheless, many of them usually contain critical information to answer a question. Therefore, they can still be used in semantic analysis for downstream tasks {\cite{allamanis2015bimodal, iyer1summarizing, zilberstein2016leveraging, yang2016query}} once paired with natural language questions}.

To systematically mine question-code pairs with high precision and recall, we propose a novel task: \textit{Given a question\footnote{Following previous work \cite{iyer1summarizing, allamanis2015bimodal, campbell2017nlp2code}, we only use the title of a question post in this work, and leave incorporating the question post content for future work.} in SO and its accepted answer post with multiple code snippets, how to predict whether each code snippet is a standalone solution or not?} In this paper, we focus on ``how-to-do-it''-type of questions which ask how to implement a certain task like in Figure~\ref{fig:runningexample}, since answers to such questions are most likely to be standalone code solutions. The definition and classification of different types of questions will be discussed in Section \ref{preliminaries}. 
We identify two challenges in our task: (1) As shown in Figure~\ref{fig:runningexample}, code snippets in an answer post can play many non-solution roles such as serving as an input-output demo or reminder (e.g., $C_2$ and $C_4$), which calls for a statistical learning model to make accurate predictions. (2) Both the textual context and the programming content of a code snippet can be predictive, but an effective model to jointly utilize them needs careful design. Intuitively, a text block with patterns like ``you can do ...'' and ``this is one thorough solution ...'' is more likely to be followed by a code solution. For example, given $S_1$ and $S_3$ in Figure \ref{fig:runningexample}, a code solution is likely to be introduced after them. On the other hand, by inspecting the code content, $C_2$ is probably not a code solution to the question, since it contains special Python console patterns like ``$>>>...>>>$'' and no particular definition of ``convert''.

To tackle these challenges, we explore a series of models including traditional classifiers and deep learning models, and propose a novel model, named Bi-View Hierarchical Neural Network (BiV-HNN), to capture both the textual context and the programming content of each code snippet (which make the two views). In BiV-HNN, we design two different modules to learn features from text and code respectively, and combine them into a deep neural network architecture, which finally predicts whether a code snippet is a standalone solution or not. 
To summarize, our contributions lie in three folds:

\textit{First}, to the best of our knowledge, we are the first to investigate \textit{systematically mining large-scale high-quality question-code pairs}, which are critical for developing learning-based models aiming to map between natural language and programming language. 

\textit{Second}, we extensively explore various models including traditional classifiers and deep learning models to predict whether a code snippet is a solution or not, and propose a novel Bi-View Hierarchical Neural Network which considers both text- and code-based views. On two manually labeled datasets in Python and SQL domain, BiV-HNN outperforms both the widely adopted heuristic methods and traditional classifiers by a large margin in terms of $F_1$ and accuracy. 
Moreover, BiV-HNN does not rely on any prior knowledge and can be easily applied to other programming domains. 

\textit{Last but not least}, we present \myData, the largest dataset to date of \textbf{$\sim$148K} Python and \textbf{$\sim$120K} SQL question-code pairs, systematically mined by our framework. Using multiple case studies, we show that (1) \myData is rich in surface variation: A question can be paired with multiple code solutions, and semantically the same code snippets can have different/paraphrased natural language descriptions. (2) Owing to such diversity as well as its large scale, \myData is a much better data resource than existing ones for constructing models to map between natural language and programming language. In addition, we can continue to grow \myData in both size and diversity, by regularly applying our framework to the fast-growing SO. Question-code pairs in other programming languages can also be mined similarly and included in \myData. 

\section{Preliminaries} \label{preliminaries}
In this section, we first clarify our task definition, and then describe how we annotated datasets for model development.

\subsection{Task Definition}

Given a question and its accepted answer post which contains multiple code snippets in Stack Overflow, we aim at predicting whether each code snippet in the answer post is a \textit{standalone} solution to the question or not. As explained in Section \ref{introduction}, we focus on ``accepted'' answer posts and ``standalone'' solutions.  

Users can ask different types of questions in SO such as ``how to implement X'' and ``what/why is Y''. Following previous work \cite{nasehi2012makes, de2014ranking, delfim2016redocumenting}, we divide questions into five types: ``How-to-do-it'', ``Debug/corrective'', ``Conceptual'', ``Seeking something, e.g., advice, tutorial'', and their combinations. In particular, a question is of type ``how-to-do-it'' when the questioner provides a scenario and asks how to implement it like in Figure \ref{fig:runningexample}. 

For collecting question-code pairs, we target at ``how-to-do-it" questions, because answers to other types of questions are not very likely to be standalone code solutions (e.g., answers to ``Conceptual'' questions are usually text descriptions). Next, we describe how to distinguish ``how-to-do-it'' questions from others. 

\subsection{``How-to-do-it'' Question Collection}
\subsubsection{Question Type Classification} \label{Qtypeclassify}~\\
At the high level, we combined the other four question types apart from ``how-to-do-it'' into one category named ``non-how-to'' and built a binary question type classifier. 

We first collected Python and SQL questions from SO based on their tags, which are available for all question posts. Specifically, we considered questions whose tags contain the keyword ``python'' to be in Python domain and questions tagged by ``sql'', ``database'' or ``oracle'' to be in SQL domain.
For each domain, we randomly sampled and labeled 250 questions for training (150), validating (20) and testing (80) the classifier\footnote{Despite of the small amount of training data, no overfitting was observed in our experiments partly because the features are very simple.}. Among the 250 questions, around 45\% in Python and 57\% in SQL are ``how-to-do-it'' questions. 
We {built} one Logistic Regression classifier respectively for each domain, based on simple features extracted from question and answer posts as in \cite{delfim2016redocumenting}, such as keyword-occurrence features, the number of code blocks in question/answer posts, the maximum length of code blocks, etc. 
Hyperparameters in classifiers were tuned based on validation sets. 

Finally, we obtained a question-type classification accuracy of 0.738 (precision: 0.653, recall: 0.889, and $F_1$: 0.753) for Python and an accuracy of 0.713 (precision: 0.625, recall: 0.946, and $F_1$: 0.753) for SQL. The classification of question types may be further improved with more advanced features and algorithms, {which is not the focus of this paper}.
\subsubsection{``How-to-do-it'' Question Set Collection}~\\
Using the above classifiers, we classified all Python and SQL questions in SO whose accepted answer post contains code blocks and collected a large set of ``how-to-do-it'' questions in each domain.
Among these ``how-to-do-it'' questions, around 44.66\% ($68,839$) Python questions and 34.45\% ($39,752$) SQL questions have an accepted answer post with more than one code snippets, from which we will systematically mine question-code pairs. 

\subsection{Annotating QC Pairs for Model Training} \label{dataset}
To construct training/validation/testing datasets for our task, we hired four undergraduate students familiar with Python and SQL to annotate answer posts in these two domains. For each code snippet in an answer post, annotators can assign ``1'' to it if they think they can solve the problem based on the code snippet alone (i.e., it is a standalone code solution), and ``0'' otherwise. We ensured each code snippet is annotated by two annotators and adopted the label only when both annotators agreed on it. {For each programming language, around 85\% code snippets were labeled.} 
The average Cohen's kappa agreement \cite{cohen1960coefficient} is around 0.658 for Python and 0.691 for SQL. The statistics of our manually {annotated} datasets are summarized in Table \ref{tab:stats}, which will be used to develop our models.  

\section{Bi-View Hierarchical NN} \label{HNN}
Without loss of generality, let us assume an answer post of a given question has a sequence of \textit{blocks} $\{S_{1}, C_{1}, S_2,..., S_i, C_i, S_{i+1}, ...,  S_{L-1}, \\C_{L-1}, S_{L}\}$ with $L$ text blocks ($S_{i}$'s) and $L-1$ code blocks ($C_{i}$'s) interleaving with each other. 
Our task is to automatically assign a binary label to each code snippet $C_i$, where 1 means a standalone solution while 0 otherwise. 
In this work, we model each code snippet independently and predict the label of $C_i$ based on its textual context (i.e., $S_i$, $S_{i+1}$) and programming content. If either $S_i$ or $S_{i+1}$ is empty, we insert an empty dummy text block to make our model applicable. One can extend our formulation to a more complicated sequence labeling problem where a sequence of code snippets can be modeled simultaneously, which we leave for future work. 

\begin{table}[t!]
\centering
\begin{tabular}{|>{\centering\arraybackslash}p{4em}|>{\centering\arraybackslash}p{3.5em}|>{\centering\arraybackslash}p{5em}|>{\centering\arraybackslash}p{3.5em}|>{\centering\arraybackslash}p{5em}|}\hline 
  \multirow{2}{3em}{} & \multicolumn{2}{c|}{\textbf{Python}} & \multicolumn{2}{c|}{\textbf{SQL}} \\ \cline{2-5}
  & \# of QC pairs & \% of QC pairs with label ``1'' & \# of QC pairs & \% of QC pairs with label ``1'' \\ \hline
  Training & 2,932 &   43.89\% & 2,183 &  56.12\% \\\hline  
  Validation & 976  & 43.14\% & 727  & 55.98\% \\\hline 
  Testing & 976  & 47.23\% & 727  & 58.32\%\\\hline 
\end{tabular}
\caption{Statistics of manually annotated datasets.} 
\vspace{-\baselineskip}
\label{tab:stats}
\end{table}

\subsection{Intuition}
\label{HNNintuition}
We first analyze at the high level how each individual block contributes to elaborating the entire answer fluently. For example, in Figure \ref{fig:runningexample}, the first text block $S_1$ suggests its followed code block $C_1$ {(which implements a function)} is ``thorough'' and thus might be a solution. $S_2$ subsequently connects $C_1$ to {examples} it can work with in $C_2$. In contrast, $S_3$ starts with the conjunction word ``Or'' and possibly will introduce an alternative solution (e.g., $C_3$). 
This observation inspires us to first model {the meaning of} each block {separately} using a \textit{token-level sequence encoder}, then model the block sequence $S_i$-$C_i$-$S_{i+1}$ using a \textit{block-level encoder}, from which we finally obtain the semantic representation of $C_i$.

\begin{figure}[t!]
\begin{center}
\includegraphics[width=0.48\textwidth]{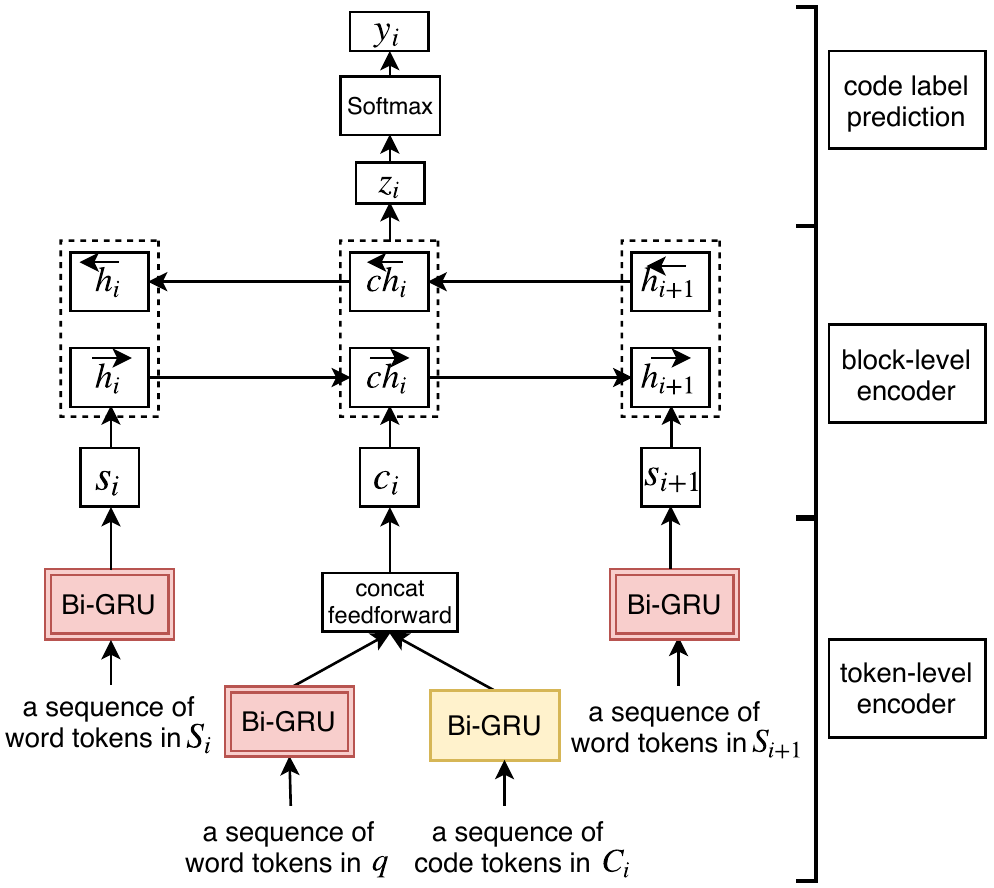}
\caption{Our Bi-View Hierarchical Neural Network (BiV-HNN). Text block $S_i$ and question $q$ are encoded by a bidirectional GRU-based RNN (Bi-GRU) module and code block $C_i$ is encoded by another Bi-GRU with different parameters.}
\label{fig:BiV-HNN}
\end{center}
\end{figure}

Figure \ref{fig:BiV-HNN} shows our model, named Bi-View Hierarchical Neural Network (BiV-HNN). It progressively learns the semantic representation of a code block from token level to block level, based on which we predict it to be a standalone solution or not. On the other hand, BiV-HNN naturally incorporates two views, i.e., textual context and code content, into the model structure.
We detail each component as follows. 

\noindent
\subsection{Token-level Sequence Encoder}\label{text-encoder}
\noindent \textbf{Text block.}
Given a sentence block $S_i$ with a sequence of words $w_{it}$, $t \in [1, T_i]$, we first embed the words into vectors through a pretrained word embedding matrix $W_e$, i.e., $x_{it} = W_e w_{it}$. We then use a bidirectional Gated Recurrent Unit (GRU) based Recurrent Neural Network (RNN) \cite{cho-al-emnlp14} to learn the word representation by summarizing the contextual information from both directions. The GRU tracks the state of sequences by controlling how much information is updated into the new hidden state from previous states. Specifically, given the input word vector $x_t$ in the current step and the hidden state $h_{t-1}$ from the last step, the GRU first computes a \textit{reset gate} $r$ for resetting information from previous steps in order to learn a new hidden state $\tilde{h}_t$: 
\begin{gather*}
r = \sigma(W_r[x_t, h_{t-1}] + b_r),\\
\tilde{h}_t = \phi(W[x_t, r \odot h_{t-1}] + b),
\end{gather*}
where $[x_t, h_{t-1}]$ is the concatenation of $x_t$ and $h_{t-1}$, $\sigma$ and $\phi$ are the \textit{sigmoid} and \textit{tanh} activation function respectively. $W_r, W$ are two weight matrices in $R^{d_h \times (d_x + d_h)}$ and $b_r, b$ are the biases in $R^{d_h}$, where $d_x, d_h$ is the dimension of $x_t$ and the hidden state $h_{t-1}$ respectively. Intuitively, if $r$ is close to 0, then the information in $h_{t-1}$ will not be passed into the current step when learning the new hidden state. The GRU also defines an \textit{update gate} $u$ for integrating hidden states $h_{t-1}$ and $\tilde{h}_t$:
\begin{gather*}
u = \sigma(W_u[x_t, h_{t-1}] + b_u),\\
h_t = uh_{t-1} + (1-u)\tilde{h}_t.
\end{gather*}
When $u$ is closer to 0, $h_t$ contains more information about the current step $\tilde{h}_t$; otherwise, it memorizes more about previous steps. Onwards, we denote the above calculation by $h_t = GRU(x_t, h_{t-1})$ for convenience.

In our work, the bidirectional GRU (i.e., Bi-GRU) contains a forward GRU reading a text block $S_i$ from $w_{i1}$ to $w_{iT_i}$ and a backward GRU which reads from $w_{iT_i}$ to $w_{i1}$:
\begin{gather*}
\overrightarrow{h}_{it} = {GRU}(x_{it}, \overrightarrow{h}_{i,t-1}), t \in [1, T_i], \\
\overleftarrow{h}_{it} = {GRU}(x_{it}, \overleftarrow{h}_{i,t+1}), t \in [T_i, 1],\\
{\overrightarrow{h}_{i0} = \overrightarrow{0}, \;\;\;\overleftarrow{h}_{i, T_i+1} = \overleftarrow{0},}
\end{gather*}
where the hidden states in both directions are initialized with zero vectors. Since the forward and backward GRU summarize the context information from different {perspectives}, we concatenate their last hidden states (i.e., $\overrightarrow{h}_{iT_i}$, $\overleftarrow{h}_{i1}$) to represent the meaning of the text block $S_i$:
$$s_i = [\overrightarrow{h}_{iT_i}, \overleftarrow{h}_{i1}].$$

\noindent\textbf{Code block.}
Similarly, we employ another Bi-GRU RNN module to learn a vector representation $v_c$ for code block $C_i$ based on its code token sequence. One may directly take this code vector $v_c$ as the token-level representation of a code block. However, since the goal of our model is to decide whether a code snippet answers a certain question, we associate $C_i$ with the question title $q$ to capture their semantic correspondences in the learnt vector representation $c_i$. Specifically, we first learn the \textit{question vector} $v_q$ by applying the token-level \textit{text} encoder to the word sequence in $q$. The concatenation of $v_q$ and $v_c$ is then fed into a feedforward \textit{tanh} layer (i.e., ``concat feedforward'' in Figure \ref{fig:BiV-HNN}) for generating $c_i$:
$$c_i = \phi (W_c[v_q, v_c] + b_c).$$
We will verify the effect of incorporating $q$ in our experiments.

Unlike modeling a code block, we do not associate a text block with question $q$ when learning its representation, because we observed no direct semantic matching between the two. For example, in Figure \ref{fig:runningexample}, a text block can hardly match the question by its content. However, as we discussed in Section \ref{introduction}, a text block with patterns like ``you can do ...'' or ``This is one {thorough solution} ...'' can imply that a code solution will be introduced after it. Therefore, we model each text block per se, without incorporating question information.

\noindent
\subsection{Block-level Sequence Encoder}
Given the sequence of token-level representations {$s_i$-$c_i$-$s_{i+1}$}, we use a bidirectional GRU-based RNN to build a block-level sequence encoder and finally obtain the code block representation: 
\begin{gather*}
\overrightarrow{h}_i = {GRU}(s_i, \overrightarrow{0}), \overleftarrow{h}_i = {GRU}(s_i, \overleftarrow{ch}_i),\\
\overrightarrow{ch}_i = {GRU}(c_i, \overrightarrow{h}_i), \overleftarrow{ch}_i = {GRU}(c_i, \overleftarrow{h}_{i+1}),\\
\overrightarrow{h}_{i+1} = {GRU}(s_{i+1}, \overrightarrow{ch}_{i}), \overleftarrow{h}_{i+1} = {GRU}(s_{i+1}, \overleftarrow{0}),
\end{gather*}
where the encoder is initialized with zero vectors (i.e., $\overrightarrow{0}$ and $\overleftarrow{0}$) in both directions. We concatenate the forward state $\overrightarrow{ch_i}$ and the backward state $\overleftarrow{ch_i}$ of the code block as its semantic representation:
$$ z_i = [\overrightarrow{ch_i}, \overleftarrow{ch_i}].$$

\noindent
\subsection{Code Label Prediction}
The representation $z_i$ of code block $C_i$ is then used for prediction:
$$y_i = \text{softmax}(W_yz_i+b_y), $$
where $y_i=[y_{i0}, y_{i1}]$ represents the probability of predicting $C_i$ to have label 0 or 1 respectively. 

We define the loss function using cross entropy \cite{Goodfellow-et-al-2016}, which is averaged over all the $N$ code snippets during training: 
$$ \mathcal{L} = - \frac{1}{N} \sum_{i=1}^{N} p_{i0}\log(y_{i0}) + p_{i1}\log(y_{i1}),$$
where $p_{i0}=0$ and $p_{i1}=1$ if the i-th code snippet is manually annotated as a solution; otherwise, $p_{i0}=1$ and $p_{i1}=0$. 

\section{Traditional Classifiers with Feature Engineering} \label{framework}
In addition to neural network based models like BiV-HNN, we also explore traditional classifiers like Logistic Regression (LR) \cite{cox1958regression} and Support Vector Machine (SVM) \cite{cortes1995support} for our task. Features are manually crafted from both text- and code-based views:

\vspace{1mm}
\noindent \textbf{Textual Context.} (1) \texttt{Token}: The unigrams and bigrams in the context. (2) \texttt{FirstToken}: If a sentence starts with phrases like ``try this'' or ``use it'', then the following code snippet is very likely to be the solution. Inspired by this idea, we discriminate the first token from others in the context. (3) \texttt{Conn}: Boolean features indicating whether a connective word/phrase (e.g., ``alternatively'') occurs in the context. We used the common connective words and phrases from Penn Discourse Tree Bank \cite{Prasad08thepenn}.

\vspace{1mm}
\noindent \textbf{Code Content.} 
(1) \texttt{CodeToken}: All code tokens in a code snippet.
(2) \texttt{CodeClass}: To discriminate code snippets that function and can be considered for learning and pragmatic reuse (i.e., ``working code'' \cite{keivanloo2014spotting}) from input-output demos, we introduce \texttt{CodeClass}, which is the probability of a code snippet being a working code. Specifically, from all the ``how-to-do-it'' Python questions in SO, we first collected totally 850 code snippets following text blocks such as ``output:'' and ``output is:'' as input-output code snippets. We further randomly selected 850 accepted answer posts containing exactly one code snippet and took their code snippets as the working code. We then extracted a set of features like the proportion of numbers and parenthesis and constructed a binary Logistic Regression classifier, which obtains 0.804 accuracy and 0.891 $F_1$ on a manually labeled testing set. Finally, the trained classifier outputs the probability for each code snippet in Python being a ``working code'' as the \texttt{CodeClass} feature. For SQL, a working code can usually be detected by keywords like ``SELECT'' and ``DELETE'', which have been included in the \texttt{CodeToken} feature. Thus, we did not design the \texttt{CodeClass} feature for it. 

\vspace{1mm}
\noindent There could be other features to incorporate into traditional classifiers. However, coming up with useful features is anything but an easy task. In contrast, neural network models can automatically learn advanced features from raw data and have been broadly and successfully applied in different areas \cite{NIPS2012_4824, simonyan2014very, mikolov2013distributed, szegedy2013deep, cho-al-emnlp14}. Therefore, in our work, we choose to design the neural network based model BiV-HNN. We will compare different models in experiments.

\section{Experiments} \label{Experiments}
In this section, we conduct extensive experiments to compare various models and show the advantages of our proposed BiV-HNN. 

\subsection{Experimental Setup}
\noindent \textbf{Dataset Summarization.} 
Section \ref{preliminaries} discussed how we manually annotated question-code pairs for training, validation and testing. Statistics were summarized in Table \ref{tab:stats}. To evaluate different models, we adopt precision, recall, $F_1$, and accuracy, which are defined in the same way as in a typical binary classification setting.

\noindent \textbf{Data Preprocessing.} We tokenized Python code snippets with best efforts: We first applied Python built-in tokenizer and for code lines that remain untokenized after that, we adopted the ``wordpunct\_tokenizer'' in NLTK toolkit \cite{nltk} to separate tokens and symbols (e.g., ``$.$'' and ``$=$''). In addition, we detected variables, numbers and strings in a code snippet by traversing its Abstract Syntax Tree (AST) parsed with Python built-in AST parser, and replaced them with special tokens ``VAR'', ``NUMBER'' and ``STRING'' respectively, to alleviate data sparsity. For SQL, we followed \cite{iyer1summarizing} to perform the tokenization, which replaced table/column names with placeholder tokens and numbered them to preserve their dependencies. Finally, we collected 4,557 (3,191) word tokens and 6,581 (1,200) code tokens from Python (SQL) training set.

\vspace{1mm}
\noindent \textbf{Implementation Details.}
We used Tensorflow\cite{tensorflow} to implement our BiV-HNN and its variants to be introduced in Section \ref{baselines}. The embedding size of word and code tokens was set at 150. The embedding vectors were pre-trained using GloVe \cite{pennington2014glove} on all Python or SQL posts in SO. Parameters were randomly initialized following \cite{glorot2010understanding}. We started the learning rate at 0.001 and trained neural network models in mini batch of size 100 with the Adam optimizer \cite{kingma2014adam}. The size of the GRU units was chosen from \{64, 128\} for token-level encoders and from \{128, 256\} for block-level encoders. Following the convention \cite{hermann2015teaching, luong-pham-manning:2015:EMNLP, iyer1summarizing}, we selected model parameters based on their performance on validation sets. The Logistic Regression and Support Vector Machine models were implemented with Python Scikit-learn library \cite{scikit-learn}.

\subsection{Baselines and Variants of BiV-HNN}\label{baselines}
\noindent \textbf{Baselines.} We compare our proposed model with two commonly used heuristics for collecting QC pairs: (1) \textit{Select-First}: Only treat the first code snippet in an answer post as a solution; (2) \textit{Select-All}: Treat every code snippet in an answer post as a solution and pair each of them with the question. In addition, we compare our model with traditional classifiers like LR and SVM based on hand-crafted features (Section \ref{framework}).

\begin{figure*}[h]
\begin{subfigure}{1.0\columnwidth}
\centering
\includegraphics[height=4.5cm, width=0.7\linewidth]{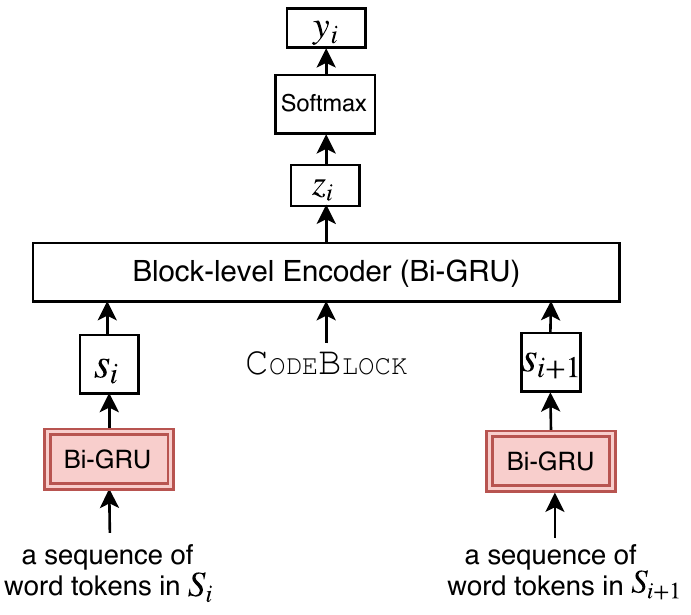} 
\caption{Text-HNN}
\label{fig:unim-hnn}
\end{subfigure}
\begin{subfigure}{1.0\columnwidth}
\centering
\includegraphics[height=4.5cm, width=0.7\linewidth]{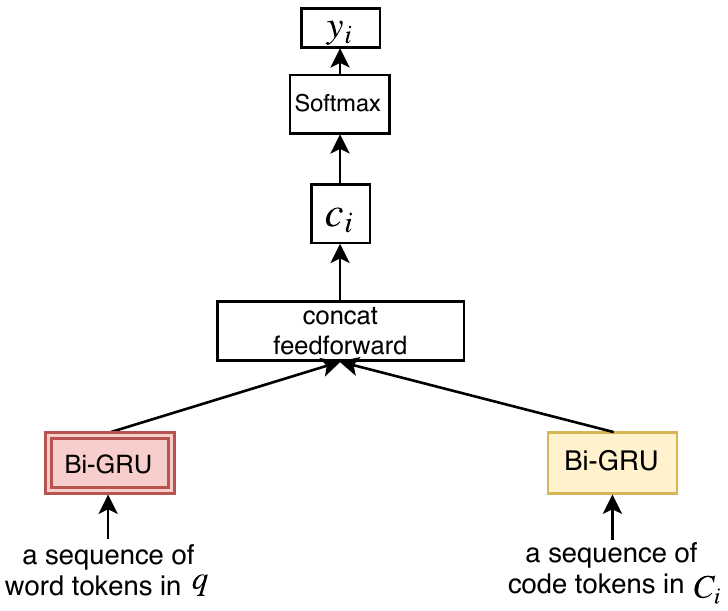} 
\caption{Code-HNN}
\label{fig:rnn-qc}
\end{subfigure}
\caption{Single-view variants of BiV-HNN: (a) Text-HNN, without code content; (b) Code-HNN, without contextual text.}
\label{fig:ablations}
\end{figure*}

\vspace{1mm}
\noindent \textbf{Variants of BiV-HNN.} \textit{First}, to evaluate the effectiveness of combining two views (i.e., textual context and code content), we adapt BiV-HNN to consider only one single view:
(1) \textit{Text-HNN} (Figure \ref{fig:unim-hnn}): In this model, we only utilize textual contexts of a code snippet. We mask all code blocks with a special token \codeblock and represent them with a unified vector. 
(2) \textit{Code-HNN} (Figure \ref{fig:rnn-qc}): We only feed the output of the token-level code encoder (i.e., $c_i$) into the ``code label prediction'' layer in Section \ref{HNN}, and do not model textual contexts.
In addition, to evaluate the effect of question $q$ when encoding a code block, we compare BiV-HNN with \textit{BiV-HNN-nq}, which directly takes the code vector $v_c$ as the code block representation $c_i$, {without associating question $q$}, for further learning. These three models are all input-level variants of BiV-HNN.

\begin{figure*}[h]
\begin{subfigure}{1.0\columnwidth}
\centering
\includegraphics[height=4.5cm]{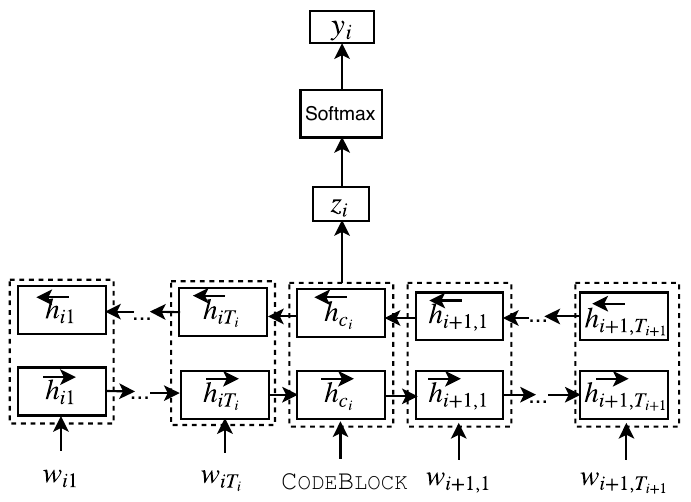} 
\caption{Text-RNN}
\label{fig:unim-rnn}
\end{subfigure}
\begin{subfigure}{1.0\columnwidth}
\centering
\includegraphics[height=4.5cm]{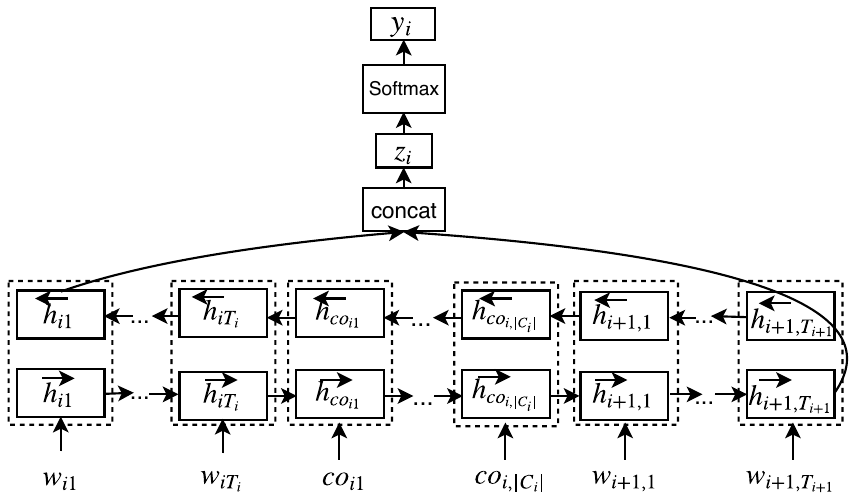} 
\caption{BiV-RNN}
\label{fig:bim-rnn}
\end{subfigure}
\caption{``Flat''-structure variants of BiV-HNN, without differentiating token- and block-level: (a) Text-RNN; (b) BiV-RNN.}
\label{fig:flat-rnn-models}
\end{figure*}

\textit{Second}, to evaluate the hierarchical structure in BiV-HNN, we compare it with ``flat'' RNN models, which model word and code tokens as a single sequence. The comparison is conducted in both text-only and bi-view settings: (1) \textit{Text-RNN} (Figure \ref{fig:unim-rnn}): Compared with Text-HNN, we concatenate all words in context blocks $S_i$ and $S_{i+1}$ as well as the unified code vector $\codeblock$ as a single sequence, i.e., $\{w_{i1},...,w_{i,T_i},\codeblock,w_{i+1,1},...,w_{i+1,T_{i+1}}\}$, using Bi-GRU RNN. The concatenation of the forward and backward hidden states of $\codeblock$ is considered as its final semantic vector $z_i${, which is then fed} into the code label prediction layer. (2) \textit{BiV-RNN} (Figure \ref{fig:bim-rnn}): In contrast to BiV-HNN, BiV-RNN models all word and code tokens in $S_i$-$C_i$-$S_{i+1}$ as a single sequence, i.e., $\{w_{i1},...,w_{iT_{i}},co_{i1}, ...,co_{ij}, ...,co_{i,|C_i|},w_{i+1,1},...,w_{i+1,T_{i+1}}\}$, where $co_{ij}$ denotes the $j$-th token in code $C_i$ and $|C_i|$ is the number of code tokens in $C_i$. BiV-RNN concatenates the last hidden states in two directions as the final semantic vector $z_i$ for prediction. {We also tried directly ``flattening'' BiV-HNN by concatenating tokens in $S_i$-$q$-$C_i$-$S_{i+1}$, but observed worse performance, perhaps because transitioning from $S_i$ to question $q$ is less natural.}

\textit{Finally}, at the block level, {instead of using an RNN}, one may apply a feedforward neural network \cite{rumelhart1988learning} to the concatenated token-level output $[s_i, c_i, s_{i+1}]$. Specifically, the block-level Bi-GRU in BiV-HNN can be replaced with a one-layer\footnote{For fair comparison, we only use one layer since the Bi-GRU in BiV-HNN only has one hidden layer.} feedforward neural network, {denoted} as \textit{BiV-HFF}. 
Intuitively, modeling the three blocks as a sequence is more consistent with the way humans read a post. We will verify this intuition in experiments.  

\vspace{1mm} 
While there could be other {variants} of our model, the above ones are related to the most critical designs in BiV-HNN. We only show their performance due to space constraints. 

\subsection{Results}
Our experimental results in Table \ref{tab:comparison-baselines} show the effectiveness of our BiV-HNN. On both datasets, BiV-HNN substantially outperforms heuristic baselines Select-First and Select-All by more than \textbf{15\%} in $F_1$ and accuracy. This demonstrates that our model can collect QC pairs with much higher quality than heuristic methods used in existing research. In addition, when compared with LR and SVM, BiV-HNN achieves $7\%$$\sim$$9\%$ higher $F_1$ and accuracy on Python dataset, and $3\%$$\sim$$5\%$ better $F_1$ and accuracy on SQL dataset. The gain on SQL data is relatively smaller, probably because {interpreting} SQL programs is a relatively easier task, implied by the observation that both simple classifiers and BiV-HNN can have around $85\%$ F1. 

\begin{table*}[ht!]
\begin{tabular}{ | c | c |c | c| c| c| c| c| c|}
  \cline{2-9}
  \multicolumn{1}{c|}{}  &\multicolumn{4}{c|}{Python Testing Set}&\multicolumn{4}{c|}{SQL Testing Set}\\ \hline
  \multicolumn{1}{|c|}{Model} & \multicolumn{1}{c|}{Precision} & \multicolumn{1}{c|}{Recall} & \multicolumn{1}{c|}{$F_1$} & \multicolumn{1}{c|}{Accuracy} & \multicolumn{1}{c|}{Precision} & \multicolumn{1}{c|}{Recall} & \multicolumn{1}{c|}{$F_1$} & \multicolumn{1}{c|}{Accuracy}\\\hline\hline
  
  \multicolumn{9}{|c|}{Heuristics Baselines}\\\hline 
  \multicolumn{1}{|c|}{Select-First} & 0.676 & 0.551 & 0.607 & 0.663 & 0.755 & 0.517 & 0.613 & 0.620 \\ 
  \multicolumn{1}{|c|}{Select-All} & 0.472 & \textbf{1.000} & 0.642 & 0.472 & 0.583 & \textbf{1.000} & 0.737 & 0.583 \\ \hline\hline
  
  \multicolumn{9}{|c|}{Classifiers based on simple features}\\ \hline
  \multicolumn{1}{|c|}{Logistic Regression} & 0.801 & 0.733 & 0.766 & 0.788 & 0.843 & 0.849  & 0.846  & 0.820 \\
  \multicolumn{1}{|c|}{Support Vector Machine}  & 0.701 & 0.813 & 0.753  & 0.748 & 0.843 & 0.858 & 0.850  & 0.824 \\ \hline\hline
  \multicolumn{1}{|c|}{\textbf{BiV-HNN}} & \textbf{0.808} & 0.876 & \textbf{0.841} & \textbf{0.843} & \textbf{0.872} & 0.903 & \textbf{0.888} & \textbf{0.867} \\ \hline
  
  \end{tabular}
  \caption{Comparison of BiV-HNN and baseline methods.}
  \label{tab:comparison-baselines}
  \vspace{-\baselineskip}
\end{table*}

\vspace{1mm}
Results in Table \ref{tab:comparison-flat} show the effect of key components in BiV-HNN in comparison with alternatives. Due to space constraints, we do not show the accuracy of each model, which has roughly the same pattern as $F_1$. We have made the following observations: 
(1) \textit{Single-view variants}. BiV-HNN outperforms Text-HNN and Code-HNN by a large margin on both datasets, showing that both views are critical for our task. In particular, by incorporating code content information, BiV-HNN is able to improve Text-HNN by 7\% on Python dataset and around 5\% on SQL dataset in $F_1$. (2) \textit{No-query variant}. On Python dataset, the integration of the question information in BiV-HNN {brings} 3\% $F_1$ improvements over BiV-HNN-nq, which shows the effectiveness of associating the question with the code snippet for identifying code answers. For SQL dataset, adding the question gives no obvious benefit, possibly because the code content in each SQL program already carries critical information for making a prediction (e.g., a SQL program containing the command keyword ``SELECT'' is very likely to be a solution to the given question, {regardless of the question content}).
(3) \textit{``Flat''-structure variants}. On both datasets, the hierarchical structure leads to $1\%$$\sim$$2\%$ improvements against the ``flat'' structure in both bi-view (BiV-HNN vs. BiV-RNN) and single-view setting (Text-HNN vs. Text-RNN). 
(4) \textit{Non-sequence variant}. On Python dataset, BiV-HNN outperforms BiV-HFF by around 2\%, showing the block-level Bi-GRU is preferable over feedforward neural networks. The two models get roughly the same performance on SQL, probably because our task is easier in SQL domain than in Python domain as we mentioned earlier. 

In summary, our BiV-HNN is much more effective than widely-adopted heuristic baselines and traditional classifiers. The key components in BiV-HNN, such as bi-view inputs, hierarchical structure and block-level sequence encoding, are also empirically justified.

\vspace{1mm}
\noindent \textbf{Error Analysis.} There are a variety of non-solution roles that a code snippet can play, such as being only one step of a multi-step solution, an input-output example, etc. We observe that more than half of the wrong predictions were false positives (i.e., predicting a non-solution code snippet as a solution), correcting which usually requires integrating information from the entire answer post. For example, when a code snippet is the first step of a multi-step solution, BiV-HNN may mistakenly take it as a complete and standalone solution, since BiV-HNN does not simultaneously take into account follow-up code snippets and their context to make predictions. In addition, BiV-HNN may make mistakes when a correct prediction requires a close examination of the \textit{content} of a question post (besides its title). Exploring these directions in the future may lead to further improved model performance on this task.

\begin{table}[t!]
\begin{tabular}{ | >{\centering\arraybackslash}p{4.5em} | >{\centering\arraybackslash}p{1.8em} |>{\centering\arraybackslash}p{1.8em} | >{\centering\arraybackslash}p{1.8em}| >{\centering\arraybackslash}p{1.8em}| >{\centering\arraybackslash}p{1.8em}| >{\centering\arraybackslash}p{1.8em}|}
  \cline{2-7}
  \multicolumn{1}{c|}{}  &\multicolumn{3}{c|}{Python Testing Set}&\multicolumn{3}{c|}{SQL Testing Set}\\ \hline
  \multicolumn{1}{|c|}{Model} & \multicolumn{1}{c|}{Prec.} & \multicolumn{1}{c|}{Rec.} & \multicolumn{1}{c|}{$F_1$} & \multicolumn{1}{c|}{Prec.} & \multicolumn{1}{c|}{Rec.} & \multicolumn{1}{c|}{$F_1$} \\\hline\hline
  
    \multicolumn{7}{|c|}{Single-view Variants}\\\hline
  \multicolumn{1}{|c|}{Text-HNN} & 0.723 & 0.826 & 0.771 & 0.798 & 0.887 & 0.840 \\
  \multicolumn{1}{|c|}{Code-HNN} & 0.770 & 0.859 & 0.812 & 0.848 & 0.854 & 0.851 \\
  \hline\hline
  
  \multicolumn{7}{|c|}{No-query Variant} \\\hline
  \multicolumn{1}{|c|}{BiV-HNN-nq} & 0.802 & 0.818 & 0.810 & \textbf{0.883} & 0.892 & 0.887 \\\hline\hline 
  
  \multicolumn{7}{|c|}{``Flat''-structure Variants}\\\hline
  \multicolumn{1}{|c|}{Text-RNN} & 0.693 & 0.824 & 0.753 & 0.773 & 0.894 & 0.829 \\ 
    \multicolumn{1}{|c|}{BiV-RNN} & 0.760 &	\multicolumn{1}{c|}{\textbf{0.887}} & 0.819 & 0.869 & 0.880 & 0.875 \\\hline\hline
    
 \multicolumn{7}{|c|}{Non-sequence Variant}\\\hline 
  \multicolumn{1}{|c|}{BiV-HFF} & 0.787 & 0.859 & 0.822 & 0.845 & \multicolumn{1}{c|}{\textbf{0.939}} & \multicolumn{1}{c|}{\textbf{0.889}}\\ \hline\hline 
  
  \multicolumn{1}{|c|}{\textbf{BiV-HNN}} & \textbf{0.808} & 0.876 & \multicolumn{1}{c|}{\textbf{0.841}} & 0.872 & 0.903 & 0.888 \\ 
  
  \hline 
  \end{tabular}
  \caption{Comparison of BiV-HNN and its variants.} 
  \label{tab:comparison-flat}
  \vspace*{-2\baselineskip}
\end{table}

\vspace{1mm}
\noindent \textbf{Model Combination.} 
When experimenting with the single-view variants of BiV-HNN, i.e., Text-HNN and Code-HNN, we observed that the three models complement each other in making accurate predictions. For example, on Python validation set, around 70\% mistakes made by Text-HNN or Code-HNN can be corrected by considering predictions from the other two models. Although BiV-HNN is built based on both text- and code-based views, 60\% of its wrong predictions can be remedied by Text-HNN and Code-HNN. The same pattern was also observed on SQL dataset.

Therefore, we further tested the effect of combining the three models via a simple heuristic: The label of a code snippet is predicted only when the three models agree on it. Using this heuristic, {69.2\%} code blocks on the annotated Python testing set are labeled with 0.916 $F_1$ and 0.911 accuracy. Similarly, on SQL testing set, 78.7\% code blocks are labeled with 0.943 $F_1$ and 0.926 accuracy. The combined model further improves BiV-HNN by around $6\%$ while still being able to label a large portion of the code snippets. Thus, we apply this combined model to those SO answer posts that are not manually annotated yet to obtain large-scale QC pairs, to be discussed next.
\section{\myData: A Systematically Mined Dataset of Question-Code Pairs} \label{mined-dataset}

In this section, we present \myData (\underline{Sta}ck Overflow \underline{Q}uestion-\underline{C}ode pairs), a large-scale and diverse set of question-code pairs automatically mined using our framework. Under various case studies, we demonstrate that \myData can greatly help tasks aiming to associate natural language with programming language.

\subsection{Statistics of \myData}
In Section \ref{Experiments}, we showed that a combination of BiV-HNN and its variants can reliably identify standalone code solutions with $>90\%$ $F_1$ and accuracy from a large portion of the testing set\nop{on testing set}. Thus we applied this combined model to all unlabeled multi-code answer posts that correspond to ``how-to-do-it'' questions in Python and SQL domain, and finally collected \textbf{60,083} and \textbf{41,826} question-code pairs respectively. 
Additionally, there are 85,294 Python and 75,637 SQL ``how-to-do-it'' questions whose answer post contains exactly one code snippet. For them, as in \cite{iyer1summarizing}, we paired the question title with the one code snippet as a question-code pair. Together with 2,169 and 2,056 manually annotated QC pairs with label ``1'' for each domain {(Table \ref{tab:stats})}, we collected a dataset of \textbf{147,546} Python and \textbf{119,519} SQL QC pairs, named as \myData. Table~\ref{tab:dataset-statistics} shows its statistics.

\vspace{0.5mm}
\textit{Note that we can continue to expand \myData with minimal efforts, since it is automatically mined by our framework, and more and more posts will be created in SO as time goes by. QC pairs in other programming languages can also be mined similarly to further enrich \myData beyond Python and SQL domain}.

\begin{table}[t]
\begin{tabular}{|>{\centering\arraybackslash}p{3.5em}|>{\centering\arraybackslash}p{3.5em}|>{\centering\arraybackslash}p{3em}|>{\centering\arraybackslash}p{3em}|>{\centering\arraybackslash}p{3em}|>{\centering\arraybackslash}p{3em}|}\hline
  \multirow{2}{4em}{} & \multirow{2}{4em}{\# of QC pairs}
  & \multicolumn{2}{c|}{\text{Question}} & \multicolumn{2}{c|}{Code} \\ \cline{3-6}
  & & Average length & \# of tokens & Average length & \# of tokens \\ \hline
  Python & \textbf{147,546} & 9 & 17,635 & 86 & 137,123 \\ \hline
  SQL & \textbf{119,519} & 9 & 9,920 & 60 & 21,143 \\ \hline 
\end{tabular}
\caption{Statistics of \myData.}
\label{tab:dataset-statistics}
\vspace*{-1.5\baselineskip}
\end{table}

\subsection{Diversity of \myData}
\begin{figure}[t]
\begin{center} 
\includegraphics[width=0.48\textwidth, height=4cm]{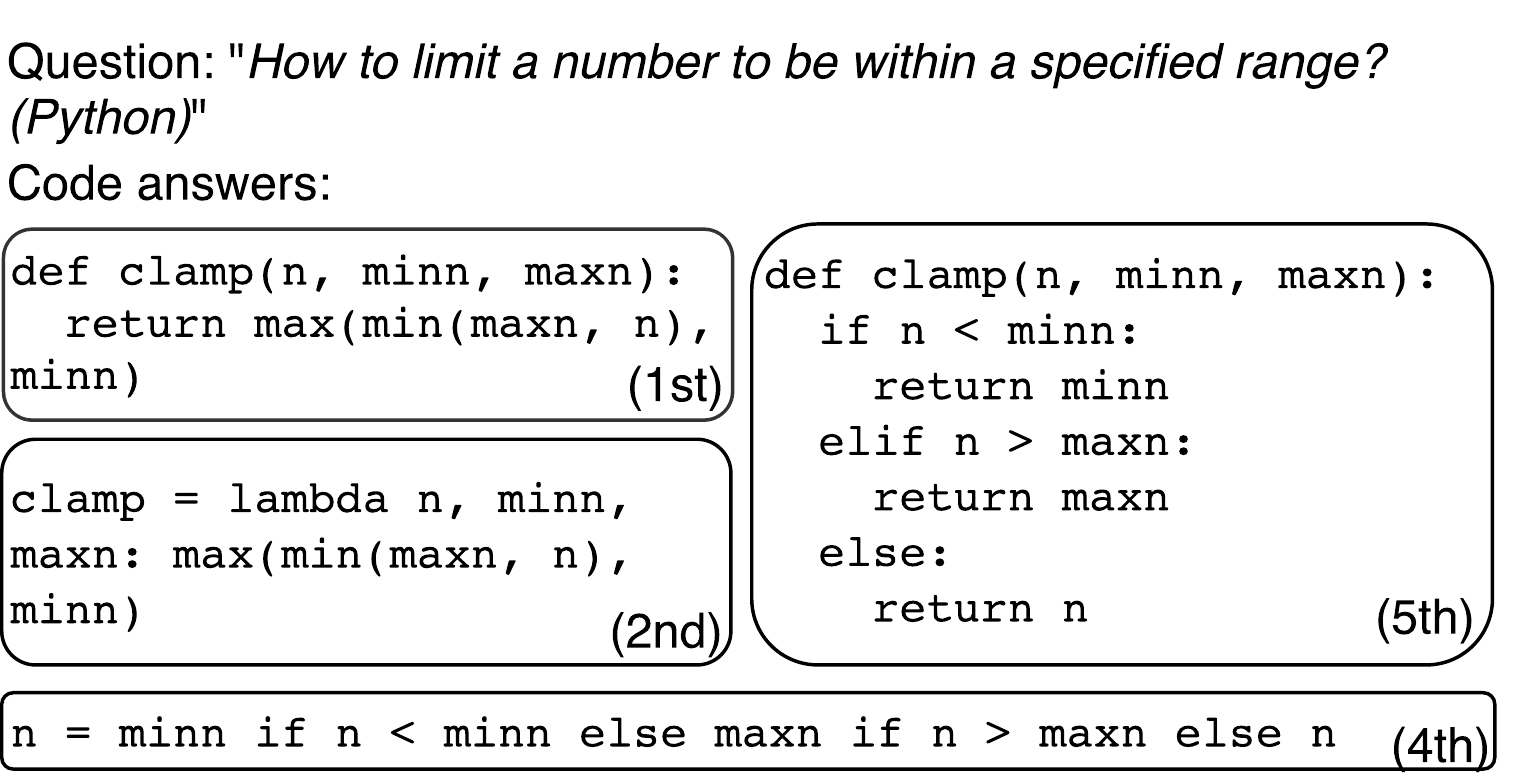} 
\caption{StaQC contains four alternative code solutions to question ``\textit{How to limit a number to be within a specified range? (Python)}'' \cite{so1} whose answer post contains five code snippets. The number at the bottom right denotes the position of each code snippet in the answer post.} 
\label{fig:case1}
\vspace*{-\baselineskip}
\end{center}
\end{figure}

\begin{figure}[t]
\begin{center}
\includegraphics[width=0.48\textwidth, height=4cm]{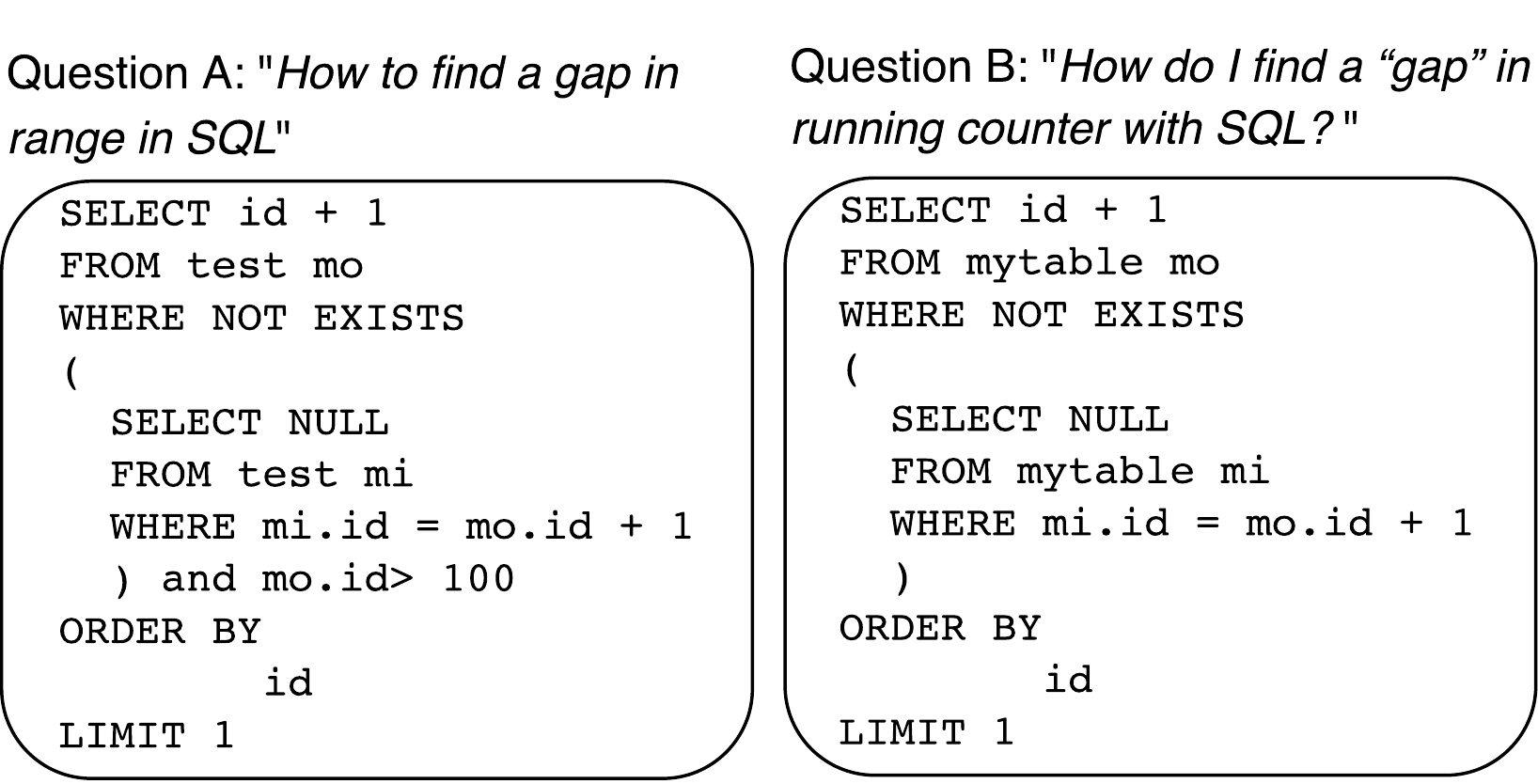} 
\caption{StaQC has different text descriptions, {e.g.,} ``\textit{How to find a gap in range in SQL}'' \cite{so2} and ``\textit{How do I find a “gap” in running counter with SQL?}'' \cite{so3}, for two code snippets bearing a similar functionality.} 
\label{fig:case2}
\vspace*{-\baselineskip}
\end{center}
\end{figure}

Besides the large scale, \myData also enjoys great diversity in the sense that it contains multiple textual descriptions for semantically similar code snippets and multiple code solutions to a question. For example, considering question ``\textit{How to limit a number to be within a specified range? (Python)}'' whose answer post contains five code snippets (Figure \ref{fig:case1}), our framework is able to correctly mine four alternative code answers. Heuristic methods may either miss some of them or mistakenly include a false solution (i.e., the 3rd code snippet). Therefore, our framework {is able to} obtain more alternative solutions for the same question more accurately. Moreover, Figure \ref{fig:case2} shows two question-code pairs included in \myData, which we easily located by comparing code solutions of \textit{relevant} questions in SO (i.e., questions manually linked by SO users). Note that the two code snippets have a very similar functionality but two different text descriptions. 

{Figure \ref{fig:case1} and \ref{fig:case2} show that \myData is highly diverse and rich in surface variation.} Such a dataset is beneficial for model development. Intuitively, when certain data patterns are not observed in the training phase, a model is less capable to predict them during testing. \myData can alleviate this issue by enabling a model to learn from alternative code solutions to the same question or from different text descriptions to similar code snippets. Next we demonstrate this benefit using an exemplar downstream task.

\subsection{Usage Demo of \myData on Code Retrieval} \label{downstream}
To further demonstrate the usage of \myData, we employ it to train a deep learning model for the code retrieval task \cite{keivanloo2014spotting, allamanis2015bimodal, iyer1summarizing}. \textit{Given a natural language description and a set of code snippet candidates, the task is to retrieve code snippets that can match the description.} In particular, an effective model should rank matched code snippets as high as possible. Models are evaluated by Mean Reciprocal Rank (MRR) \cite{voorhees1999trec}. In \cite{iyer1summarizing}, the authors proposed a neural network based model, \textit{CODE-NN}, which outputs a matching score between a natural language question and a code snippet. We choose CODE-NN as it is one of the state of the arts for {code retrieval} and improved previous work by a large margin. For training, the authors collected {around 25,870 SQL QC pairs from answer posts containing exactly one code snippet (which is paired with the question title)}. They manually annotated two datasets DEV and EVAL for choosing the best model parameters and for final evaluation respectively, both containing around 100 QC pairs. The final evaluation is conducted in 20 runs. In each run, for every QC pair in DEV or EVAL, \cite{iyer1summarizing} randomly selected 49 code snippets from SO as non-answer candidates, and ranked all 50 code snippets based on their scores output by CODE-NN. The averaged MRR is computed as the final result. 

\vspace{1mm}
\noindent \textbf{Improved Retrieval Performance.} 
We first trained CODE-NN using the \textit{original} training set in \cite{iyer1summarizing}. We denote this setting as \textit{CODE-NN (original)}. Then we used \myData to upgrade the training data in two most straightforward ways: (1) We directly took all the 119,519 SQL QC pairs in \myData to train CODE-NN, denoted as \textit{CODE-NN (\myData)}. (2) To emphasize the effect of our framework, we just added the \text{41,826} QC pairs, automatically mined from SO multi-code answer posts, to the original training set and retrained the model, which is denoted as \textit{CODE-NN (original + \myData-multi)}. In both (1) and (2), {questions and code snippets} occurring in the {DEV/EVAL} set were removed from training. 

{In all three settings, we used the same DEV/EVAL set and the same hyper-parameters as in \cite{iyer1summarizing} except the dropout rate, which was chosen from \{0.5, 0.7\} for each model to obtain better performance.} Like \cite{iyer1summarizing}, we decayed the learning rate in each epoch and terminated the training when it was lower than 0.001. The best model was selected as the one achieving the highest {average} MRR on DEV set. When using this strategy, we observed better results on the EVAL set than those reported in \cite{iyer1summarizing} (around 0.44).

Table \ref{tab:downstream-comparison} shows the average MRR score and standard deviation of each model on EVAL set. We can see that directly using StaQC for training leads to a substantial 6\% improvement over using the original dataset in \cite{iyer1summarizing}. By adding QC pairs we mined from multi-code posts to the original training data, CODE-NN can be significantly improved by 3\%. Note that the performance gains shown here are still conservative, since we adopted the same hyper-parameters and a small evaluation set, in order to see the direct impact of \myData. 
Using more challenging evaluation sets and by conducting systematic hyper-parameter selection, we expect models trained on \myData to be more advantageous. \myData can also be used to train other code retrieval models besides CODE-NN, as well as models for other related tasks like code generation or annotation. 

\begin{table}[t]
\begin{tabular}[width=0.5\textwidth]{|c|c|}\hline
  Model Setting & MRR \\\hline 
  CODE-NN (original) & 0.51 $\pm$ 0.02 \\ \hline 
  \textbf{CODE-NN (\myData)}$^*$ & \textbf{0.57} $\pm$ \textbf{0.02} \\\hline 
  CODE-NN (original + \myData-multi)$^*$ & 0.54 $\pm$ 0.02 \\ \hline
 \end{tabular}
\caption{Performance of CODE-NN \cite{iyer1summarizing} on code retrieval, with and without \myData for training. $^*$denotes {statistically significant} w.r.t. \textit{CODE-NN (original)} under one-tailed Student's t-test ($p<0.05$).}
\label{tab:downstream-comparison}
\vspace*{-1.5\baselineskip}
\end{table}
\section{Discussion and Future Work}
Besides boosting relevant tasks using \myData, future work includes: 
(1) We currently only consider a code snippet to be a standalone solution or not. In many cases, code snippets in an answer post serve as multiple steps and should be merged to form a complete solution \cite{so4}. This is a more challenging task and we leave it to the future. 
(2) In our experiments, we combined BiV-HNN and its two variants using a simple heuristic to achieve better performance. In the future, one can also use \myData to retrain the three models, similar to self-training \cite{nigam2000analyzing}, or jointly train the three models in a tri-training framework \cite{zhou2005tri}.
(3) One may also employ Convolutional Neural Networks \cite{shen2014latent, NIPS2012_4824, allamanis2016convolutional}, which have shown great power on representation learning, to encode text and code blocks. Moreover, we can consider encoders similar to \cite{nguyen2015graph, mou2016convolutional} for capturing the intrinsic structure of programming language.
\section{Related Work}

\textbf{\textit{Language} + \textit{Code} Tasks and Datasets}. 
Tasks that map between natural language and programming language, referred to \textit{Language} + \textit{Code} tasks here, such as code annotation and code retrieval/generation, have been popularly studied in recent years \cite{giordani2009semantic, keivanloo2014spotting, oda2015learning, allamanis2015bimodal, iyer1summarizing, raghothaman2016swim, zilberstein2016leveraging, ling2016latent, vinayakarao2017anne}. In order to train more advanced yet data-hungry models, researchers have collected data either automatically from online communities \cite{keivanloo2014spotting, allamanis2015bimodal, iyer1summarizing, zilberstein2016leveraging, raghothaman2016swim, ling2016latent, vinayakarao2017anne, barone2017parallel} or with human intervention \cite{giordani2010corpora, oda2015learning}. Like our work, \cite{allamanis2015bimodal, iyer1summarizing, zilberstein2016leveraging, vinayakarao2017anne} utilized SO to collect data. Particularly, \cite{allamanis2015bimodal} merges code snippets in its answer post as the target source code and pair it with the question title. \cite{iyer1summarizing} only employs accepted answer posts containing exactly one code snippet. Other interesting datasets include $\sim$19K $<$English pseudo-code, Python code snippet$>$ pairs manually annotated by \cite{oda2015learning}, and $\sim$114K pairs of Python functions and their documentation strings heuristically collected by \cite{barone2017parallel} from GitHub \cite{github}.
Unlike their work, we systematically mine high-quality question-code pairs from SO using advanced machine learning models. Our mined dataset \myData, the largest to date of around 148K Python and 120K SQL question-code pairs, has been shown to be a better resource. Moreover, \myData is easily expandable in terms of both scale and programming language types.

\vspace{0.5mm}
\noindent
\textbf{Recurrent Neural Networks for Sequential Data}.
Recurrent Neural Networks have shown great success in various natural language tasks \cite{bahdanau2014neural, cho-al-emnlp14, luong-pham-manning:2015:EMNLP, hermann2015teaching}. In an RNN, terms are modeled sequentially without discrimination. Recently, in order to handle information at different levels, \cite{li2015hierarchical, serban2016building, tang2015document, yang2016hierarchical} stack multiple RNNs into a hierarchical structure. 
For example, \cite{yang2016hierarchical} incorporates the attention mechanism in a hierarchical RNN model to pick up important words and sentences. Their model finally aggregates all sentence vectors to learn the document representation. In comparison, we utilize the hierarchical structure to first learn the semantic meaning of each block individually, and then predict the label of a code snippet by {combining two views: textual context and programming content.}

\vspace{0.5mm}
\noindent
\textbf{Mining Stack Overflow}. Stack Overflow has been the focus of the Mining Software Repositories (MSR) challenge for years \cite{MSRChallenge2013, MSRChallenge2015}. A lot of work \cite{treude2011programmers, nasehi2012makes, de2014ranking, duijn2015quality, yang2016query, delfim2016redocumenting} have been done on exploring the categories of questions, mining source codes, etc. We follow \cite{nasehi2012makes, de2014ranking, delfim2016redocumenting} to categorize SO questions into 5 classes but only focus on the ``how-to-do-it'' type (Section \ref{preliminaries}). \cite{duijn2015quality, yang2016query} analyzes the quality of code snippets (e.g., readability) or explores ``usable'' code snippets that could be parsed, compiled and run. Different from their work, we are interested in finding standalone code solutions, which are not necessarily directly parsable, compilable or runnable, but can be semantically paired with questions\nop{(e.g., even if they are pseudo-code)}. To the best of our knowledge, we are the first to study the problem of systematically mining high-quality question-code pairs.
\section{Conclusion}
This paper explores systematically mining question-code pairs from Stack Overflow, {in contrast to heuristically collecting them}. 
We focus on the ``how-to-do-it'' questions since their answers are more likely to be code solutions. We present the largest-to-date dataset of diversified question-code pairs in Python and SQL domain (\myData), systematically collected by our framework. \myData can greatly help downstream tasks aiming to associate natural language with programming language. We will release it together with our source code for future research.

\section*{Acknowledgments}
This research was sponsored in part by the Army Research Office under cooperative agreements W911NF-17-1-0412, Fujitsu gift grant, DARPA contract FA8750-13-2-0019, the University of Washington WRF/Cable Professorship, Ohio Supercomputer Center \cite{OhioSupercomputerCenter1987}, and NSF Grant CNS-1513120. The views and conclusions contained herein are those of the authors and should not be interpreted as representing the official policies, either expressed or implied, of the Army Research Office or the U.S. Government. The U.S. Government is authorized to reproduce and distribute reprints for Government purposes notwithstanding any copyright notice herein. 

\newpage
\balance

\bibliographystyle{ACM-Reference-Format}


\end{document}